\documentclass[10pt,twocolumn,letterpaper]{article}

\usepackage{cvpr}
\usepackage{times}
\usepackage{epsfig}
\usepackage{graphicx}
\usepackage{amsmath}
\usepackage{amssymb}
\usepackage{algorithmic}
\usepackage{algorithm}
\usepackage{enumitem}

\usepackage[breaklinks=true,bookmarks=false]{hyperref}

\cvprfinalcopy 

\def\cvprPaperID{****} 

\begin{document}

\title{Automatic Curation of Golf Highlights using Multimodal Excitement Features\vspace{-0.3cm}}

\author{Michele Merler$^1$
\and
Dhiraj Joshi$^1$
\and
Quoc-Bao Nguyen$^1$
\and
Stephen Hammer$^2$
\and
John Kent$^2$ \vspace{0.1cm}
\and
\rule{0.35in}{0pt} John R. Smith$^1$ 
\and
Rogerio S. Feris$^1$ 
\and 
\\  $^1$IBM T. J. Watson Research Center
\hspace{0.5in}
$^2$ IBM iX 
}

\maketitle

\renewcommand\UrlFont{\color{blue}\rmfamily\itshape}

\begin{abstract}
The production of sports highlight packages summarizing a game's most exciting moments is an essential task for broadcast media. Yet, it requires labor-intensive video editing. We propose a novel approach for auto-curating sports highlights, and use it to create a real-world system for the editorial aid of golf highlight reels. Our method fuses information from the players' reactions (action recognition such as high-fives and fist pumps), spectators (crowd cheering), and commentator (tone of the voice and word analysis) to determine the most interesting moments of a game. We accurately identify the start and end frames of key shot highlights with additional metadata,  such as the player's name and the hole number, allowing personalized content summarization and retrieval. In addition, we introduce new techniques for learning our classifiers with reduced manual training data annotation by exploiting the correlation of different modalities. Our work has been demonstrated at a major golf tournament, successfully extracting highlights from live video streams over four consecutive days. 
\end{abstract}


\vspace{-0.15in}
\section{Introduction} \label{sec:intro}

The tremendous growth of video data has resulted in a significant demand for tools that can accelerate and simplify the production of sports highlight packages for more effective browsing, searching, and content summarization. In a major professional golf tournament such as Masters, for example, with 90 golfers playing multiple rounds over four days, video from every tee, every hole and multiple camera angles can quickly add up to hundreds of hours of footage. Yet, most of the process for producing highlight reels is still manual, labor-intensive, and not scalable.

\begin{figure}[t]
\begin{center}
\includegraphics[width=0.98\linewidth]{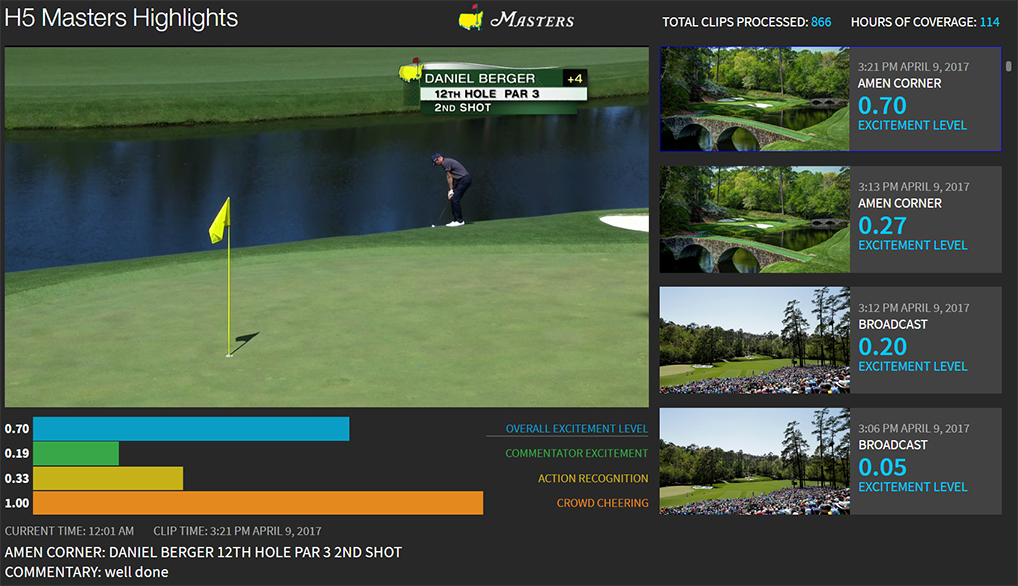}
\end{center}
\vspace{-0.3cm}
   \caption{The H5 system dashboard for auto-curation of sports highlights. Highlights are identified in near real-time (shown in the right panel) with an associated excitement level score. The user can click on the icons in the right panel to play the associated video in the center, along with the scores for each excitement measure. }
\label{fig:short}
\end{figure}

In this paper, we present a novel approach for auto-curating sports highlights, showcasing its application in extracting golf play highlights. Our approach uniquely fuses information from the {\em player}, {\em spectators}, and the {\em commentator} to determine a game's  most exciting moments. More specifically, we measure the excitement level of video segments based on the following multimodal markers:
\vspace{-0.2cm}
\begin{itemize}[noitemsep]
\item {\bf Player reaction:} visual action recognition of player's celebration (such as high fives or fist pumps);
\item {\bf Spectators:} audio measurement of crowd cheers;
\item {\bf Commentator: } excitement measure based on the commentator's tone of the voice, as well as exciting words or expressions used, such as ``beautiful shot''.
\end{itemize}

\vspace{-0.2cm}
These indicators are used along with the detection of TV graphics (e.g., lower third banners) and shot-boundary detection to accurately identify the start and end frames of key shot highlights with an overall excitement score. The selected segments are then added to an interactive dashboard for quick review and retrieval by a video editor or broadcast producer, speeding up the process by which these highlights can then be shared with fans eager to see the latest action. Figure \ref{fig:short} shows the interface of our system, called High-Five ({\bf High}lights {\bf F}rom {\bf I}ntelligent {\bf V}ideo {\bf E}ngine), H5 in short.

In our approach, we exploit how one modality can guide the learning of another modality, with the goal of reducing the cost of manual training data annotation. In particular, we show that we can use TV graphics and OCR as a proxy to build rich feature representations for player recognition from {\em unlabeled} video, without requiring costly training data annotation. Our audio-based classifiers also rely on feature representations learned from unlabeled video \cite{soundnetNIPS16}, and are used to constrain the training data collection of other modalities (e.g., we use the crowd cheer detector to select training data for player reaction recognition). 
Personalized highlight extraction and retrieval is another unique feature of our system. By leveraging TV graphics and OCR, our method automatically gathers information about the player's name and the hole number. This metadata is matched with the relevant highlight segments, enabling searches like ``show me all highlights of player X at hole Y during the tournament'' and personalized highlights generation based on a viewer's favorite players. 
In summary, the key {\bf contributions} of our work are listed below:
\vspace{-0.2cm}
\begin{itemize}[noitemsep]
\item We present a first-of-kind system for automatically extracting golf highlights by uniquely fusing multimodal excitement measures from the player, spectators, and commentator. In addition, by automatically extracting metadata via TV graphics and OCR, we allow personalized highlight retrieval or alerts based on player name, hole number, location, and time.

\item Novel techniques are introduced for learning our multimodal classifiers without requiring costly manual training data annotation. In particular, we build rich feature representations for player recognition without manually annotated training examples. 

\item We provide an extensive evaluation of our work, showing the importance of each component in our proposed approach, and comparing our results with professionally curated highlights. Our system has been successfully demonstrated at a major golf tournament, processing live streams and extracting highlights from four channels during four consecutive days.

\end{itemize}
\vspace{-0.2cm}

\begin{figure*}[ht]
\begin{center}
\includegraphics[width=0.88\linewidth]{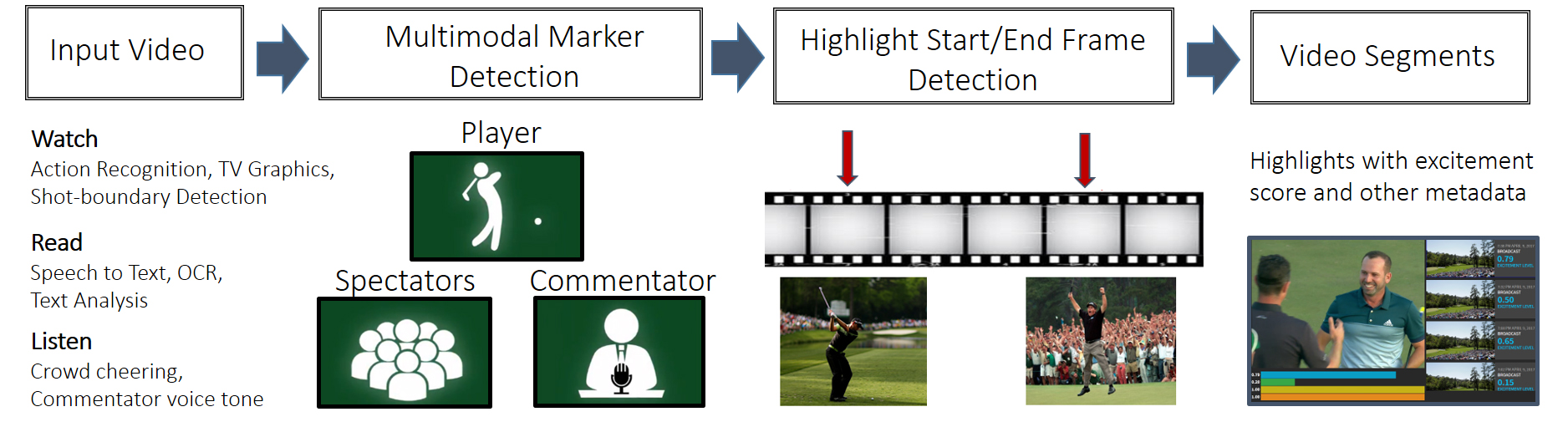}
\end{center}
\caption{Our approach consists of applying multimodal (video, audio, text) marker detectors to measure the excitement levels of the player, spectators, and commentator in video segment proposals. The start/end frames of key shot highlights are accurately identified based on these markers, along with the detection of TV graphics and visual shot boundaries. The output highlight segments are associated with an overall excitement score as well as additional metadata such as the player name, hole number, shot information, location, and time. }
\label{fig:framework}
\end{figure*}

\section{Related Work} \label{sec:related}
\vspace{0.05in}
{\bf Video Summarization.} There is a long history of research on video summarization \cite{ma2002user, rav2006making,zhang2016video}, which aims to produce short videos or keyframes that summarize the main content of long full-length videos. Our work also aims at summarizing video content, but instead of optimizing for representativeness and diversity as traditional video summarization methods, our goal is to find the highlights or exciting moments in the videos.  A few recent methods address the problem of highlight detection in consumer videos \cite{sun2014ranking,yang2015unsupervised,yao2016highlight}. Instead our focus is on sports videos, which offer more structure and more objective metrics than unconstrained consumer videos.

\vspace{0.05in}
{\bf Sports Highlights Generation.} Several methods have been proposed to automatically extract highlights from sports videos based on audio and visual cues. Example approaches include the analysis of replays \cite{zhao2006highlight}, crowd cheering \cite{xiong2003audio}, motion features \cite{xiong2003generation}, and closed captioning \cite{zhang2002event}. More recently, Bettadapura et al. \cite{bettadapura2016leveraging} used contextual cues from the environment to understand the excitement levels within a basketball game. Tang and Boring \cite{tang2012epicplay} proposed to automatically produce highlights by analyzing social media services such as twitter. Decroos et al. \cite{decroos2017predicting} developed a method for forecasting sports highlights to achieve more effective coverage of multiple games happening at the same time. Different from existing methods, our proposed approach offers a unique combination of excitement measures to produce highlights, including information from the {\em spectators}, the {\em commentator}, and the {\em player} reaction. In addition, we enable personalized highlight generation or retrieval based on a viewer`s favorite players.

\vspace{0.05in}
{\bf Self-Supervised Learning.} In recent years, there has been significant interest in methods that learn deep neural network classifiers without requiring a large amount of manually annotated training examples. In particular, {\em self-supervised} learning approaches rely on auxiliary tasks for feature learning, leveraging sources of supervision that are usually available ``for free'' and in large quantities to regularize deep neural network models. Examples of auxiliary tasks include the prediction of ego-motion \cite{agrawal2015learning,jayaraman2015learning}, location and weather \cite{wang2016walk}, spatial context or patch layout \cite{noroozi2016unsupervised,pathak2016context}, image colorization \cite{zhang2016colorful}, and temporal coherency \cite{mobahi2009deep}. Aytar et al. \cite{soundnetNIPS16} explored the natural synchronization between vision and sound to learn an acoustic representation from {\em unlabeled} video. We leverage this work to build audio models for crowd cheering and commentator excitement with a few training examples, and use these classifiers to constrain the training data collection for player reaction recognition. More interestingly, we exploit the detection of TV graphics as a free supervisory signal to learn feature representations for player recognition from unlabeled video.

\section{Technical Approach} \label{sec:system}

\subsection{Framework}
Our framework is illustrated in Figure \ref{fig:framework}. Given an input video feed, we extract in parallel four multimodal markers of potential interest: player action of celebration (detected by a visual classifier), crowd cheer (with an audio classifier), commentator excitement (detected by a combination of an audio classifier and a salient keywords extractor applied after a speech-to-text component). We employ the audience cheer detector for seeding a potential moment of interest. Our system then computes shot boundaries for that segment as exemplified in Figure \ref{fig:startend}. The start of the segment is identified by graphic content overlaid to the video feed. By applying an OCR engine to the graphic, we can recognize the name of the player involved and the hole number, as well as additional metadata. The end of the segment is identified with standard visual shot boundary detection applied in a window of few seconds after the occurrence of the last excitement marker. 
Finally we compute a combined excitement score for the segment proposal based on a combination of the individual markers. 
In the following we describe each component in detail.


\subsection{Audio-based Markers Detection}  \label{ssec:audio}
Crowd cheering is perhaps the most veritable form of approval of a player's shot within the context of any sport. Specifically in golf, we have observed that cheers almost always accompany important shots. Most importantly crowd cheer can point to the fact that an important shot was just played (indicating the end of a highlight). Another important audio marker is excitement in the commentators' tone while describing a shot. Together those two audio markers play an important role in determining the position and excitement level of a potential highlight clip. 

In this work, we leverage Soundnet \cite{soundnetNIPS16} to construct audio-based classifiers for both crowd-cheering and commentator tone excitement. Soundnet uses a deep 1-D convolutional neural network architecture to learn representations of environmental sounds from nearly 2 million unlabeled videos. Specifically, we extract features from the {\it conv-5} layer to represent 6 seconds audio segments. The choice of the {\it conv-5} layer is based upon experiments and superior results reported in \cite{soundnetNIPS16}. The dimensionality of the feature is 17,152. 
One key advantage of using such a rich representation pre-trained on millions of environmental sounds is the direct ability to build powerful linear classifiers, similarly to what has been observed for image classification \cite{razavian2014baseline}, for cheer and commentator tone excitement detection with relatively few audio training examples (for example we started with 28 positive and 57 negative training samples for the audio-based commentator excitement classifier). 
We adopt an iterative refinement bootstrapping methodology to construct our audio based classifiers. We learn an initial  classifier with relatively few audio snippets and then perform bootstrapping on a distinct test set. This procedure is repeated to improve the accuracy at each iteration. 
\vspace{-0.2cm}
\subsubsection{Crowd Cheer Detection} \label{ssec:cheer}
Cheer samples from 2016 Masters replay videos as well as examples of cheer obtained from YouTube were used in order to train the audio cheer classifier using a linear SVM on top of deep features. For negative examples, we used audio tracks containing regular speech, music, and other kinds of non-cheer sounds found in Masters replays.
In total our final training set consisted of 156 positive and 193 negative samples (6 seconds each). The leave-one-out cross validation accuracy on the training set was 99.4\%. 
\vspace{-0.2cm}
\subsubsection{Commentator Excitement Detection}\label{ssec:tone}
We propose a novel commentator excitement measure based on voice tone and speech-to-text-analysis.
{\bf Tone-based:} Besides recognizing crowd cheer, we employ the deep Soundnet audio features to model excitement in commentators' tone. As above, we employ a linear SVM classifier for modeling. For negative examples, we used audio tracks containing regular speech, music, regular cheer (without commentator excitement) and other kinds of sounds which do not have an excited commentator found in 2016 Masters replays. In total, the training set for audio based commentator excitement recognition consisted of 131 positive and 217 negative samples. The leave-one-out cross validation accuracy  on the training set was 81.3\%. \linebreak
{\bf Text-based:} While the commentator's tone can say a lot about how excited they are while describing a particular shot, the level of their excitement can also be gauged from another source, that is, the expressions they use. We created a dictionary of 60 expressions (words and phrases) indicative of excitement (e.g. "great shot", "fantastic" ) and assign to each of them excitement scores ranging from 0 and 1. We use a speech to text service\footnote{\url{https://www.ibm.com/watson/developercloud/speech-to-text.html}} to obtain a transcript of commentators' speech and create an excitement score as an aggregate of scores of individual expressions in it.

\begin{figure*}[ht]
\begin{center}
\includegraphics[width=0.8\linewidth]{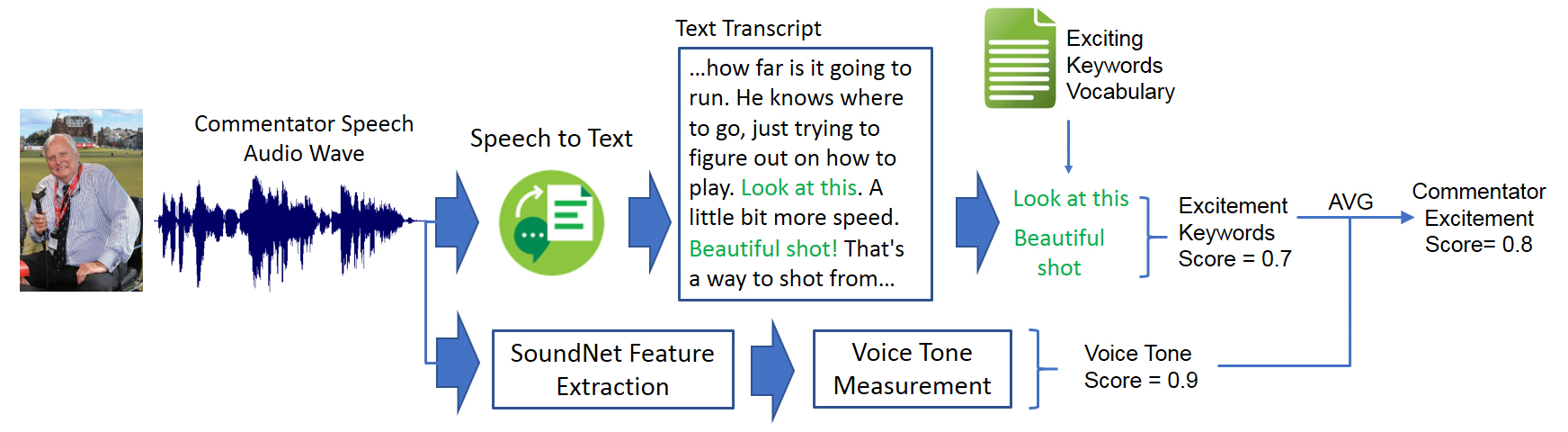}
\end{center}
\vspace{-0.3cm}
\caption{Commentator excitement score computation based on (i) audio tone analysis and (ii) speech to text analysis.}
\label{fig:commentator}
\end{figure*}

When assigning a final excitement score to a highlight (as described in Section~\ref{ssec:fusion}), we average the tone-based and text-based commentator excitement to obtain the overall level of excitement of the commentator. The two scores obtained from complementary sources of information create a robust measure of commentator excitement, as exemplified in  Figure~\ref{fig:commentator}. 


\subsection{Visual Marker Detection}  \label{ssec:visual}

\subsubsection{Player Reaction} \label{ssec:action}
Understanding the reaction of a player is another important cue to determine an interesting moment of a game. In our work, we train an action recognizer to detect a player celebrating. To the best of our knowledge, measuring excitement from the player reaction for highlight extraction has not been explored in previous work.

\begin{figure}
\begin{center}
\includegraphics[width=0.8\linewidth]{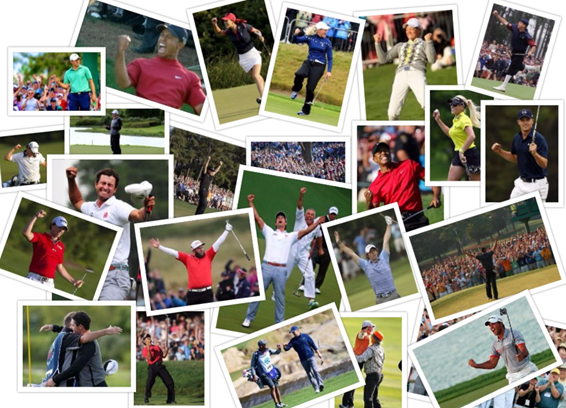}
\end{center}
\vspace{-0.3cm}
\caption{Examples of still images used to train our action recognition model.}
\label{fig:action}
\end{figure}

We adopt two strategies to reduce the cost of training data collection and annotation for action recognition. First, we use our audio-based classifiers (crowd cheer and commentator excitement) at a low threshold to select a subset of video segments for annotation, as in most cases the player celebration is accompanied by crowd cheer and/or commentator excitement. Second, inspired by \cite{ma2017less}, we use still images which are much easier to annotate and allow training with less computational resources compared to video-based classifiers. Figure \ref{fig:action} shows examples of images used to train our model. At test time, the classifier is applied at every frame and the scores aggregated for the highlight segment as described in the next Section. 

Initially, we trained a classifier with 574 positive examples and 563 negative examples. The positive examples were sampled from 2016 Masters replay videos and also from the web. The negative examples were randomly sampled from the Masters videos. We used the
VGG-16 model \cite{simonyan2014very}, pre-trained on Imagenet as our base model. The Caffe \cite{jia2014caffe} deep learning library was used to train the model with stochastic gradient descent, learning rate 0.001, momentum 0.9, weight decay 0.0005. Then, we performed three rounds of hard negative mining on Masters videos from previous years, obtaining 2,906 positive examples and 6,744 negative ones. The classifier fine-tuned on this data achieved  88\% accuracy on a separate test set containing 460 positive and 858 negative images.

\subsubsection{TV Graphics, OCR, and Shot-boundaries} \label{ssec:shot}
In professional golf tournament broadcasts, a golf swing is generally preceded by a TV graphics with the name of the player just about to hit the golf ball and other information about the shot. The detection of such markers is straightforward, as they appear in specific locations of the image, and have distinct colors. We check for such colors in the vicinity of pre-defined image locations (which are fixed across all broadcasting channels) to determine the TV graphics bounding box. One could use a more general approach by training a TV graphics detector (for example via faster-rcnn \cite{renNIPS15fasterrcnn} or SSD \cite{liu2016ssd}), however this was beyond the scope of this work. We then apply OCR (using the Tesseract engine \cite{smith2007overview}) within the detected region in order to extract metadata such as the name of the player and the hole number. This information is associated with the detected highlights, allowing personalized queries and highlight generation based on a viewer’s favorite players. We also use standard shot-boundary detection based on color histograms \cite{amir2003ibm} as a visual marker to better determine the end of a highlight clip.

\subsection{Highlight Detection} \label{ssec:fusion}
\begin{figure*}
\begin{center}
\includegraphics[width=0.77\linewidth]{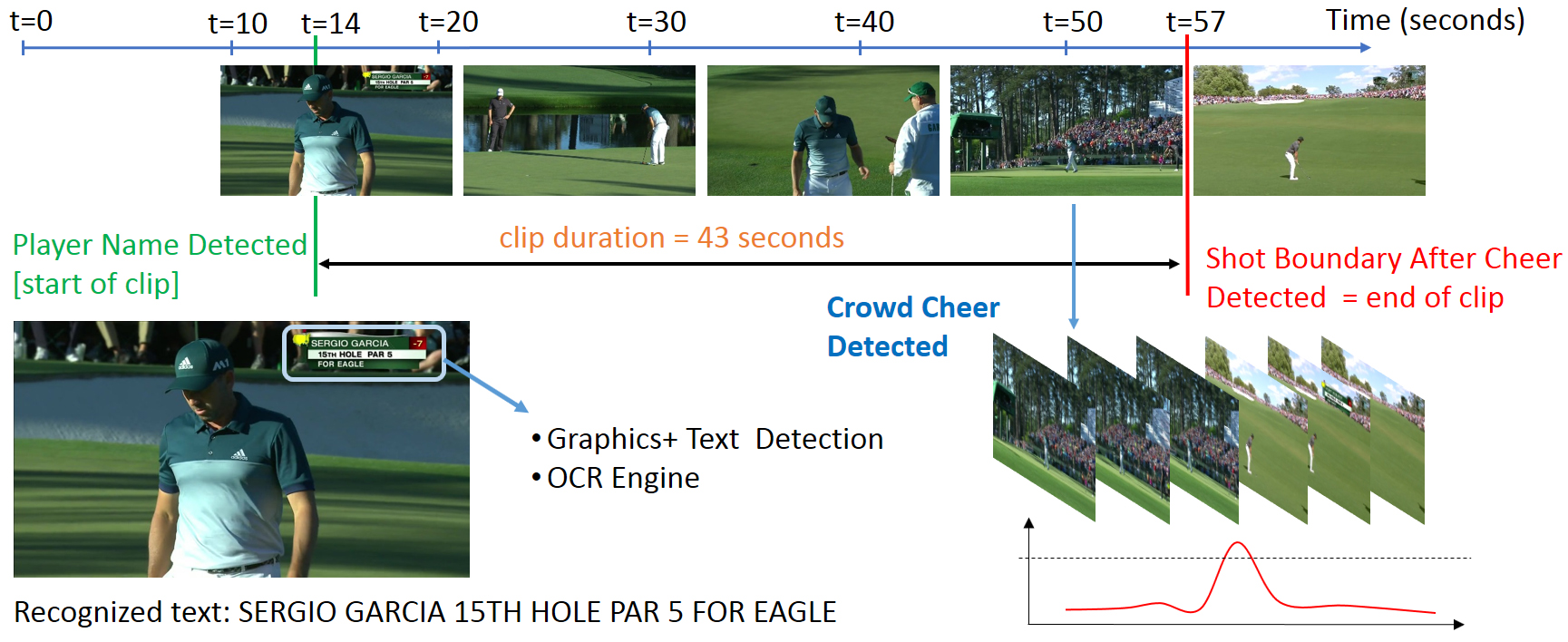}
\end{center}
\vspace{-0.3cm}
\caption{Demonstrating highlight clip start and end frames selection.}
\label{fig:startend}
\end{figure*}

Figure \ref{fig:startend} illustrates how we incorporate multimodal markers to identify segments as potential highlights and assign excitement scores to them. The system starts by generating {\bf segment proposals} based on the crowd cheering marker. Specifically, crowd cheering detection is performed on a continuous segment of the stream and positive scores are tapped to point to potentially important cheers in audio. Adjacent 6 second segments with positive scores are merged to mark the end of a bout of contiguous crowd cheer. Each distinct cheer marker is then evaluated as a potential candidate for a highlight using presence of a TV graphics marker containing a player name and hole number within a preset duration threshold (set at 80 seconds). The beginning of the highlight is set as 5 seconds before the appearance of TV graphics marker. In order to determine the end of the clip we perform shot boundary detection in a 5 second video segment starting from the end of cheer marker. If a shot boundary is detected, the end of the segment is set at the shot change point.

Segments thus obtained constitute valid highlight segment proposals for the system. The highest cheer score value among adjacent segments that are merged is set as the crowd cheer marker score for a particular segment proposal. Once those baseline segment scores have been computed, we perform further search to determine if the segment contains player celebration action, excitement in commentators' tone, or exciting words or expressions used to describe the shot. It is important to note that the cheer and commentator excitement predictions are performed on every 6 seconds audio segment tapped from the video stream. Similarly the visual player celebration action recognition is performed on frames sampled at 1 fps.   

In order to determine the overall excitement level of a video segment we incorporate available evidence from all audio, visual, and text based classifiers that fall within a segment proposal. Specifically, we aggregate and normalize positive scores for these markers within a time-window of detected crowd cheer marker. For player reaction, we set this window to be 15 seconds while for audio commentator excitement the window was set to be 20 seconds. 
Finally we obtain the overall excitement score of a segment proposal using a linear fusion of scores obtained from crowd cheer, commentator excitement (audio and text-based), and player celebration action markers. 
Weights for crowd cheer, commentator excitement (audio and text) and player reaction components are set as 0.61, 0.13, 0.13, and 0.13 respectively. The search time-windows, segment duration thresholds and weights for linear fusion were decided on the basis of analysis performed on the training set, which consists on the broadcast from the 2016 Masters tournament.


\section{Self-Supervised Player Recognition} \label{sec:zeroshot}

Automatic player detection and recognition can be a very powerful tool for generating personalized highlights when graphics are not available, as well as to perform analysis outside of the event broadcast itself. It could for example enable to estimate the presence of a player in social media posts by recognizing his face. The task is however quite challenging. First, there is a large variations in pose, illumination, resolution, occlusion (hats, sunglasses) and facial expressions, even for the same player, as visible in Figure \ref{fig:face}. Second, inter-player differences are limited, as many players wear extremely similar outfits, in particular hats, which occlude or obscure part of their face. Finally, a robust face recognition model requires large quantities of labeled data in order to achieve high levels of accuracy, which is often difficult to obtain and labor intensive to annotate. We propose to alleviate such limitations by exploiting the information provided by other modalities of the video content, specifically the overlaid graphics containing the players name. This allows us to generate a large set of training examples for each player, which can be used to train a face recognition classifier, or learn powerful face descriptors.

We start by detecting faces within temporal window after a graphic with a player name is found, using a faster-rcnn detector \cite{renNIPS15fasterrcnn}. The assumption is that in the segment after the name of a player is displayed, his face will be visible multiple times in the video feed. Not all detected faces in that time window are going to represent the player of interest. We therefore perform  outliers removal, using geometrical and clustering constraints. We assume the distribution of all detected faces to be bi-modal, with the largest cluster containing faces of the player of interest. Faces that are too small are discarded, and faces in a central position of the frame are given preference. Each face region is expanded by 40\% and rescaled to 224x224 pixels. Furthermore, only a maximum of one face per frame can belong to a given player. Given all the face candidates for a given player, we perform two-class k-means clustering on top of fc7 features extracted from a VGG Face network \cite{Parkhi15}, and keep only the faces belonging to the largest cluster while respecting the geometric constraints to be the representative examples of the player's face.
This process, working without supervision, allows us to collect a large quantity of training images for each player. We can then train a player face recognition model,  which in our case consists of a VGG Face Network fine-tuned by adding a softmax layer with one dimension per player. Figure \ref{fig:face}(b) shows an example subset of training faces automatically collected for Sergio Garcia from the 2016 Golf Masters broadcast. The system was able to collect hundreds of images with a large variety of pose and expressions for the same player. Bordered in red are highlighted two noisy examples. While the purity of the training clusters is not perfect, as we will show in the experiments of Section \ref{ssec:resultface} it still allowed to learn a robust classifier with no explicit supervision.

\section{Experiments} \label{sec:experiments}

\subsection{Experimental Setting} \label{ssec:setting}
We evaluated our system in a real world application, namely the 2017 Golf Masters tournament. We analyzed in near real-time the content of the four channels broadcasting simultaneously over the course of four consecutive days, from April 6th to April 9th, for a total of 124 hours of content\footnote{Video replays are publicly available at \url{http://www.masters.com/en_US/watch/index.html}}. Our system produced 741 highlights over all channels and days.
The system ran on a Redhat Linux box with two K40 GPUs. We extracted frames directly from the video stream at a rate of 1fps and audio in 6 seconds segments encoded as 16bit PCM at rate 22,050. The cheer detector and commentator excitement run in real time (1 second to process one second of content), the action detection takes 0.05secs per frame, graphics detection with OCR takes 0.02secs per frame. The speech-to-text is the only component slower than real time, processing 6 seconds of content in 8 seconds, since we had to upload every audio chunk to an API service.
In the following we report experiments conducted after the event to quantitatively evaluate the performance of the system, both in terms of overall quality of the produced highlights as well as the efficacy of its individual components.
All training was performed on content from the 2016 Golf Masters broadcast, while testing was done on the last day of the 2017 tournament.


\subsection{Highlights Detection} \label{ssec:resulthighlights}

Evaluating the quality of sports highlights is a challenging task, since a clearly defined ground truth does not exist. Similarly to previous works in this field \cite{bettadapura2016leveraging}, we approached this problem by comparing the clips automatically generated by our system to two human based references. The first is a human evaluation and ranking of the clips that we produced. The second is the collection of highlights professionally produced by the official Masters curators and published on their Twitter channel.

\subsubsection{Human Evaluation of Highlights Ranking}
\begin{figure}[t]
\begin{center}
\includegraphics[width=0.9\linewidth]{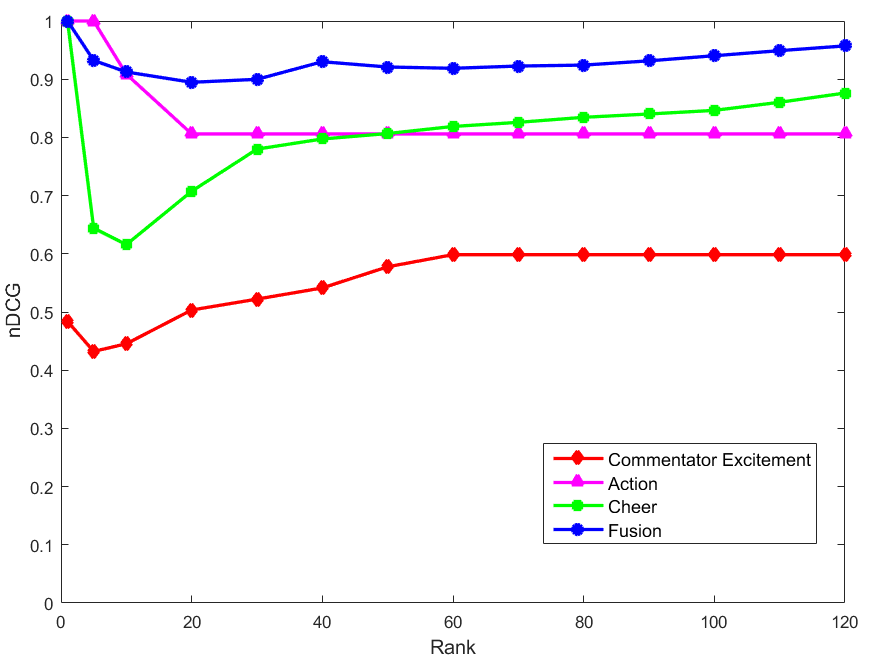}
\end{center}
\caption{nDGC computed at different ranks for the individual components as well as the Fusion.}
\vspace{-0.3cm}
\label{fig:nDCG}
\end{figure}

We employed three persons in a user study to determine the quality of the top 120 highlights clips produced by our system from Day 4 of the Golf Masters. We asked each participant to assign a score to every clip in a scale from 0 to 5, with 0 meaning a clip without any interesting content, 1 meaning a highlight that is associated with the wrong player, and 2 to 5 meaning true highlights, 5 being the most exciting shots and 2 the least exciting (but still relevant) shots. We then averaged the scores of the three users for each clip.
The resulting scores determined that 92.68\% of the clips produced by our system were legitimate highlights (scores 2 and above), while 7.32\% were mistakes. We also compared the rankings of the clips according to the scores of each individual component, as well as their fusion, to the ranking obtained through the users votes. The performance of each ranking is computed at different depth $k$ with the normalized discounted cumulative gain (nDCG) metric, which is a standard retrieval measure computed as follows
\[
nDCG(k) = \frac{1}{Z} \sum^k_{i=1} \frac{2^{rel_i} -1}{log_2(i+1)}
\]
where $rel_i$ is the relevance score assigned by the users to clip $i$ and $Z$ is a normalization factor ensuring that the perfect ranking produces a nDCG score of 1.
In Figure \ref{fig:nDCG} we present the nDGC at different ranks. We notice that all components but the Commentator Excitement correctly identify the most exciting clip (at rank 1). After that only the Action component assigns the highest scores to the following top 5 clips. When considering 10 top clips or more, the benefit of combining multiple modalities becomes apparent, as the Fusion nDGC curve remains constantly higher than each individual marker. 

\subsubsection{Comparison with Official Masters Highlights}
The previous experiment confirmed the quality of the identified highlights as perceived by potential users of the system. We then compared H5 generated clips with highlights professionally created for Masters, {\it Masters Moments}, available at their official Twitter page\footnote{\url{https://twitter.com/mastersmoments}}.
There are a total of 116 highlight videos from the final day at the 2017 Masters. Each one covers a player's approach to a certain hole (e.g. Daniel Berger, 13th hole) and usually contains multiple shots that the player used to complete a particular hole. In contrast each H5 highlight video is about a specific shot at a particular hole for a given player. 
 In order to match the two sets of videos, we considered just the player names and hole numbers and ignored the shot numbers. After eliminating Masters Moments outside of the four channels we covered live during the tournament and for which there is no matching player graphics marker, we obtained 90 Masters Moments. 

In Table~\ref{tab:highlightsresults}, we report Precision and Recall of matching clips over the top 120 highlights produced by the H5 Fusion system. We observe that approximately half of the clips overlap with Masters Moments. This leaves us with three sets of videos: one shared among the two sets (a gold standard of sorts), one unique to Masters Moments and one unique to H5. 
We observed that by lowering thresholds on our markers detectors, we can incorporate 90\% of the Masters Moments by producing more clips. Our system is therefore potentially capable of producing almost all of the professionally produced content.
We also wanted to investigate the quality of the clips which were discovered by the H5 system beyond what the official Master's channel produced. 
Generation of highlights is a subjective task and may not comprehensively cover every player and every shot at the Masters. At the same time, some of the shots included in the official highlights may not necessarily be great ones but strategically important in some ways. 

While our previous experiment was aimed at understanding the coverage of our system vis-a-vis official Masters highlights, we wondered if a golf aficionado would find the remaining videos still interesting (though not part of official highlights).
We therefore aimed an experiment at quantitatively comparing (a) H5 highlight clips that matched Masters Moments and (b) H5 highlight clips that did not match Masters Moments videos. 

In order to do so we selected the 40 most highly ranked (by H5) videos from lists (a) and (b) respectively and performed a user study using three human participants familiar with golf. Participants were shown pairs of videos with roughly equivalent H5 scores/ranks (one from list (a) and the other from list (b) above) and were asked to label the more interesting video between the two, or report that they were equivalent. Majority voting was used among the users votes to determine the video pick from each pair. From the results reported in Table \ref{tab:highlightsresults} we observe that while the preference of the users lies slightly more for videos in set (a), in almost half of the cases the highlights uniquely and originally produced by the H5 system were deemed equally if not more interesting. This reflects that the system was able to discover content that users find interesting and goes beyond what was officially produced.
It is also interesting to notice that our system is agnostic with respect to the actual score action of a given play, that is, a highlight is detected even when the ball does not end up in the hole, but the shot is recognized as valuable by the crowd and/or commentator and players through their reactions to it.


\begin{table}[t]
\centering
\begin{tabular}{|c|c|c|}
\hline
Depth & 120 & 500  \\ \hline \hline
Precision & 0.54  & 0.35 \\ \hline
Recall & 0.4 & 0.9 \\ \hline \hline
Matching Highlights Preference & 0.57 & - \\ \hline
Non-Matching Highlights Preference & 0.33 & - \\ \hline
Equivalent & 0.10 & - \\ \hline
\hline
\end{tabular}
\vspace{0.2cm}
\caption{Highlights detection performance. Comparison between the top $k$ ($k=120,500$) retrieved clips from our system and the official Master's Twitter highlights.}
\label{tab:highlightsresults}
\end{table}



\subsection{Self-Supervised Recognition} \label{ssec:resultface}

\begin{figure}[t]
\begin{center}
\includegraphics[width=\linewidth]{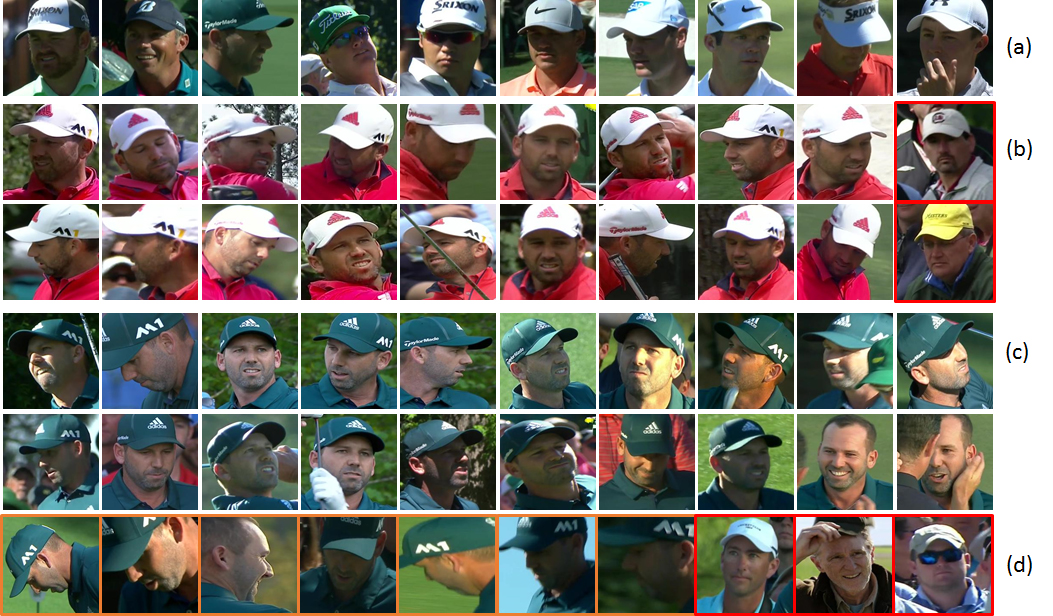}
\end{center}
\vspace{-0.3cm}
\caption{Self-supervised player face learning. (a) Examples of the 10 players used in the experiments. (b) Subset of the images automatically selected as training set (2016 Masters) for Sergio Garcia (note the diversity of pose, expression, occlusion, illumination, resolution). (c) Examples of test faces (2017 Masters) correctly recognized through self-supervised learning. (d) Examples of False Negatives (in orange) and False Positives (in red).}
\label{fig:face}
\end{figure}

In order to test our self-supervised player recognition model we randomly selected a set of 10 players who participated to both the 2016 and the 2017 tournaments (shown in Figure \ref{fig:face} (a)). 
In Table \ref{tab:faceresults} we report the statistics of the number of training images that the system was able to automatically obtain in a self-supervised manner. For each player we obtain on average 280 images. Data augmentation in the form of random cropping and scaling was performed to uniform the distribution of examples across players. Since there is no supervision in the training data collection process, some noise in bound to arise. We manually inspected the purity of each training cluster (where one cluster is the set of images representing one player) and found it to be 94.26\% on average. Note that despite evaluating its presence, we did not correct for the training noise, since our method is fully self-supervised.
The face recognition model was fine-tuned from a face VGG network with learning rate = 0.001, $\gamma = 0.1$, momentum = 0.9 and weight decay = 0.0005. The net converged after approximately 4K iterations with batch size 32.
We evaluated the performance of the model on a set of images randomly sampled from Day 4 of the 2017 tournament and manually annotated with the identity of the 10 investigated players.
Applying the classifier directly to the images achieved 66.47\% accuracy (note that random guess is 10\% in this case since we have 10 classes). We exploited the fact that the images come from video data to cluster temporally close images based on fc7 features and assigned to all images in a cluster the identity which received the highest number of predictions within the cluster. This process raised the performance to 81.12\%.
Figure \ref{fig:face} (c) shows examples of correctly labeled test images of Sergio Garcia. Note the large variety of pose, illumination, occlusion and facial expressions. In row (d) we also show some examples of false negatives (bordered in orange) and false positives (in red).
The net result of our framework is thus a self-supervised data-collection procedure which allows to gather large quantities of training data without need for any annotation, which can be used to learn robust feature representations and face recognition models.

\begin{table}[t]
\centering
\begin{tabular}{|c|c|c|c|}
\hline
 Number of Players & 10\\ \hline \hline
 Number of Training Images & 2,806\\ \hline
 Training Clusters Purity & 94.26\%\\ \hline \hline
 Number of Test Images & 1,181\\ \hline 
 Random Guess &  10.00\%\\  \hline
 Classifier Alone Accuracy & 66.47\%\\  \hline
 Classifier + Clustering Accuracy & \textbf{81.12\%}\\ \hline
\hline
\end{tabular}
\vspace{0.2cm}
\caption{Face classification performance. }
\label{tab:faceresults}
\end{table}

\subsection{Discussion}
While we have demonstrated our approach in golf, we believe our proposed techniques for modeling the excitement levels of the players, commentator, and spectators are general and can be extended to other sports. The way we determine the start of an event based on TV graphics is specific to golf, but that could be replaced by other markers in other sports. In tennis, for example, the start of an event could be obtained based on the detection of a serve by action recognition. 

The combination of multimodal excitement measures is crucial to determine the most exciting moments of a game. Crowd cheer is the most important marker, but alone cannot differentiate a hole-in-one or the final shot of the tournament from other equally loud events. In addition, we noticed several edge cases where non-exciting video segments had loud cheering from other holes. Our system correctly attenuates the highlight scores in such cases, due to the lack of player celebration and commentator excitement. We believe that other sources of excitement measures, such as player and crowd facial expressions, or information from social media could further enhance our system.

The same approach used for self-supervised player recognition could also be applied for the detection of golf setup (player ready to hit the golf ball), using TV graphics as a proxy to crop positive examples based on person detection. This would generalize our approach to detect the start of an event without relying on TV graphics, and also help fix a few failure cases of consecutive shots for which a single TV graphics is present.

\section{Conclusion} \label{sec:conclusion}
We presented a novel approach for automatically extracting highlights from sports videos based on multimodal excitement measures, including audio analysis from the spectators and the commentator, and visual analysis of the players. Based on that, we developed a first-of-a-kind system for auto-curation of golf highlight packages, which was demonstrated in a major tournament, accurately extracting the start and end frames of key shot highlights over four days. We also exploited the correlation of different modalities to learn models with reduced cost in training data annotation. As next steps, we plan to demonstrate our approach in other sports such as tennis and produce more complex storytelling video summaries of the games.

\small
{
\bibliographystyle{ieee}
\bibliography{CVsports17}
}

\end{document}